\pdfoutput=1
\documentclass[11pt]{article}

\usepackage[]{acl}

\usepackage{times}
\usepackage{latexsym}

\usepackage[T1]{fontenc}
\usepackage[utf8]{inputenc}

\usepackage{microtype}
\usepackage{wrapfig}

\usepackage{array}
\usepackage{boldline}
\usepackage{graphicx}
\usepackage{booktabs}

\usepackage{multirow}

\title{Leveraging Expert Guided Adversarial Augmentation \\For Improving Generalization in Named Entity Recognition}

\author{Aaron Reich$^{1,2}$, Jiaao Chen$^1$, Aastha Agrawal$^1$, Yanzhe Zhang$^1$, Diyi Yang$^1$\\
   $^1$Georgia Institute of Technology\\
   $^2$Pionetechs, Inc.\\
  \texttt{\{areich8, jchen896, aagrawal319, z\_yanzhe, dyang888\}@gatech.edu}}

\begin{document}
\maketitle
\begin{abstract}
Named Entity Recognition (NER) systems often demonstrate great performance on in-distribution data, but perform poorly on examples drawn from a shifted distribution. 
One way to evaluate the generalization ability of NER models is to use adversarial examples, on which the specific variations associated with named entities are rarely considered. To this end, we propose leveraging expert-guided heuristics to change the entity tokens and their surrounding contexts thereby altering their entity types as adversarial attacks. Using expert-guided heuristics, we augmented the CoNLL 2003 test set and manually annotated it to construct a high-quality challenging set. We found that state-of-the-art NER systems trained on CoNLL 2003 training data drop performance dramatically on our challenging set. By training on adversarial augmented training examples and using mixup for regularization, we were able to significantly improve the performance on the challenging set as well as improve out-of-domain generalization which we evaluated by using OntoNotes data. We have publicly released our dataset and code at \url{https://github.com/GT-SALT/Guided-Adversarial-Augmentation}. 
\end{abstract}

\section{Introduction}

Deep learning models have achieved great performance on many natural language processing (NLP) problems \cite{bahdanau2016neural, devlin-etal-2019-bert}. However, many recent works have shown that these models often rely on \textit{spurious correlations} which are not necessarily the \textit{causal artifacts}. Thus, these models perform well on the in-distribution test set but are likely to exhibit a huge performance decline on out-of-distribution data (e.g. real world data) \cite{DBLP:journals/corr/abs-2007-06778, kaushik-lipton-2018-much, poliak2018hypothesis, gururangan-etal-2018-annotation, DBLP:journals/corr/abs-1904-01130, glockner2018breaking}. Prior works have constructed adversarial examples for benchmarking the generalization ability of state-of-the-art NLP models on out-of-distribution examples \cite{kaushik2020learning, DBLP:journals/corr/abs-1904-01130, glockner2018breaking}.
% The adversarial examples that do involve high linguistic quality augmentations are often human annotated such as in \cite{kaushik2020learning}.
%\diyi{would suggest combining the above two paragraphs - they are good, but take very slow steps and too long to open the story. the key is adversial examples for the next paragraph.}
Proposed approaches such as random word swapping \cite{jin2020bert} and the appending of a sentence to the end of text \cite{jia2017adversarial} do not take into consideration the unique linguistic properties and variations associated with named entities.
As a key problem setting involving the classification of semantic categories of entities (e.g., Organizations, Locations)  \cite{jbp:/content/journals/10.1075/li.30.1.03nad}, NER is still in need of improved benchmarks of true generalization.

Previous works \cite{hardeval, fu2020rethinking, stanislawek-etal-2019-named} have shown that 
% \textit{overlapping}, \textit{ambiguous} categories or \textit{label shift} where a 
words which have different entity labels in different scenarios often lead to frequently occurring errors of NER models. 
This can be especially problematic in specific domain applications where this challenging case is common. For example, when training an NER model for political text mining, it would be of great importance to differentiate between the categories of \textsl{Clinton} (Person) and the \textsl{Clinton Foundation} (Organization). We make use of this as the inspiration for designing expert-guided heuristic linguistic patterns for creating a high quality adversarial dataset for NER.

Leveraging such expert-guided heuristics, we propose an automated procedure for adversarial augmentation. We use this automated procedure to first generate adversarial examples from the test data. 
Since some of these automatically generated adversarial examples may lack quality in terms of syntax or semantics, we manually select only the examples that are of high quality for the construction of the challenging test set. The performance of state-of-the-art NER systems drops severely on this challenging test set.
To alleviate this degradation, we first use the proposed heuristics to augment the training examples (without manually filtering the data for quality), which proves to be effective.
We further utilize mixup \cite{zhang2018mixup, chen2020local} as a regularization technique to interpolate the representations of the original examples and the augmented examples, leading to a smoother decision boundary and improved generalization ability \citep{lee2020adversarial, wang2021augmax}.

\section{Related Work}
%\diyi{do you think your related work and introduction can be combined? using space till 132 lines should be reasonable.} I think the way the intro and related work are divided works. 
\paragraph{Generating Adversarial Examples} Adversarial data augmentation \citep{chen2021empirical} severely influences a model's predictions without changing human judgements. It is widely leveraged to test the generalization ability of models \citep{wang2021adversarial}.
For example, \citet{jia2017adversarial} fools a reading comprehension system by inserting distracting sentences. \citet{belinkov2018synthetic} leverages synthesized or natural typos to attack character-based translation models.
% Adversarial GLUE \citep{wang2021adversarial} employed the use of expert-annotated templates for constructing adversarial data, However, it does not include NER as a task. 
However, few prior works have explored the generation of adversarial examples specifically for NER. \citet{gui2021textflint} performed augmentations by concatenating sentences, swapping/inserting/deleting a random character in an entity, entity swapping with Out-of-Vocabulary entities, and cross category swapping. \citet{zeng-etal-2020-counterfactual} also took a random entity swapping approach but only selected entities of the same label to preserve linguistic correctness. In this work, we purposely alter the entity type by adding/deleting tokens in predefined \emph{word phrase} sets and alter the surrounding context.

\paragraph{Adversarial Training and Mixup}
One approach for improving a model's performance on adversarial examples is to incorporate adversarial examples into its training (adversarial training, \citealp{goodfellow2014explaining}). However, this may not improve the generalization ability of the model, since the model is only learning to focus on manipulated hard examples \citep{lee2020adversarial}.
One solution is to combine mixup \citet{zhang2018mixup} with adversarial training \citep{lee2020adversarial, wang2021augmax}. By linearly interpolating training data and their associated labels, mixup is able to improve the classifier's generalization ability by training on these interpolated data points which helps to form a smoother decision surface. In the context of adversarial training, mixup is leveraged to form diverse adversarial examples \citep{wang2021augmax} and prevent overfitting on adversarial features \citep{lee2020adversarial}, thus improving the overall generalization ability.
% Several previous studies \cite{thulasidasan2020mixup, verma2019manifold} showed how using mixup allows neural models to be less confident during distribution shifts than models not using mixup. It also leads to a smoother decision-making surface than Empirical Risk Minimization (ERM), thereby making models more robust to adversarial examples \cite{8478159}. % As mixup generates new latent data points through linear interpolation of original data, it alleviates the dependency on human-annotated data \cite{chen2020local, chen2020mixtext}, especially in low resource settings \cite{sun2020mixuptransformer}. %
In this work, we use mixup to interpolate the original examples and expert-guided adversarial examples to improve the generalization ability of NER models.
%\diyi{can you add this paper: \url{https://openreview.net/forum?id=GF9cSKI3A_q}and compare how your work relates to or differs from it?} 
\section{Expert-Guided Adversary Generation}
\label{sec:advgen}

\begin{table*}[ht]
\centering
\small
\begin{tabular}{p{0.15\linewidth} | p{0.05\linewidth} | p{0.70\linewidth}}
\hline
\textbf{Transition} & \textbf{Count} & \textbf{Examples}\\
\hline
\multirow{2}{*}{\begin{tabular}[c]{@{}l@{}}Location or Person \\ $\rightarrow$ Organization\end{tabular}}                              & \multirow{2}{*}{510}       & \textbf{Original:} Every year, 500 new plastic surgeons graduate in \color{blue}Brazil \color{black} and medical students from all over the world come to study there. \\
            &                      & \textbf{Augmented:}Every year, 500 new plastic surgeons graduate from \color{red} Brazil University \color{black} and	medical	students from all over	the world come	to study there. \\
                                                                                             \hline

\begin{tabular}[c]{@{}l@{}}Organization $\rightarrow$\\ Location\end{tabular}                               & 99       & \begin{tabular}[c]{@{}l@{}}\textbf{Original:} \color{blue} Munich Re \color{black} says to split stock.\\ \textbf{Augmented:} \color{red} Munich\color{black}\color{brown}'s largest corporation \color{black} says to split stock.\end{tabular}\\
                                                                                             \hline
           \multirow{2}{*}{\begin{tabular}[c]{@{}l@{}}Organization or \\ Location $\rightarrow$ Person\end{tabular}}                    & \multirow{2}{*}{391}       & \textbf{Original:} The \color{blue}Colts \color{black} won despite the absence of injured starting defensive tackle Tony Siragusa, cornerback Ray Buchanan and linebacker Quentin Coryatt. \\
            &                      & \textbf{Augmented:} \color{red} Colts Zardari \color{brown} and her team \color{black} won despite the absence of injured starting defensive tackle Tony Siragusa, cornerback Ray Buchanan and linebacker Quentin Coryatt. \\
                                                                                             \hline

\end{tabular}
\caption{\small{Expert-guided transition types for producing adversarial augmentations for NER. The original entity is colored in \color{blue} blue \color{black} and entity token change is colored in \color{red} red\color{black}. The entity context change is colored in \color{brown} brown\color{black}. Note that the entity context change is not always applied in the transition to \textsc{Organization}. We also provided the statistics of the challenging set.}}
\label{table:heuristic_rule_based_augmentation}
\end{table*}
Current NER models often deal with unambiguous cases where one entity often gets assigned to the same label. By inducing challenging cases using the Overlapping Categories \cite{fu2020rethinking} that alter the entity and its label, models can then be tested to see whether they are only learning spurious correlations between the token and the label.
For the construction of adversarial examples by the altering of entity types, we define three components: (i) \textbf{Eligibility Check}: We only augment entities that are eligible to change their entity types. (ii) \textbf{Entity Token Change}: By adding or deleting certain predefined tokens, we change the entity type of the original tokens to a target type. (iii) \textbf{Entity Context Change}: To deal with ambiguous tokens, we further add some predefined contexts that correspond to the target entity type. Note that predefined words/phrases/contexts used in different scenarios form different predefined \emph{word phrase} sets, into which embed expert knowledge. During the automatic generation process, we randomly sample from the corresponding \emph{word phrase} sets. Table \ref{table:heuristic_rule_based_augmentation}  
contains examples of expert-guided adversarial augmentations. The three components are defined below for their use in the transition to each target entity type (organization, person, location):

\paragraph{Organization} For transitioning to \textsc{Organization}, an example is considered \textbf{eligible} if an entity only contains one token (e.g. ``\emph{Brazil}''). \textbf{Entity Token Change} in this case refers to inserting words and phrases which are often used behind or after some tokens to form an organization (e.g. add ``\emph{University}'' after ``\emph{Brazil}''). Such words and phrases form a set of size 44, including ``\emph{University of}'' (inserted before) and ``\emph{Department}'' (inserted after). \textbf{Entity Context Change} for \textsc{Organization} involves inserting a suitable context after the newly formed organization entity, such as ``\emph{and its team}'' and ``\emph{'s office}''. Such phrases form a set of size 42.

\paragraph{Location} Different from transitioning to \textsc{Organization}, we want to instead ensure the augmented entity of type \textsc{Location} is a real world location. To achieve this, we \textbf{combine} the eligibility check and entity token change: we first define a \emph{word phrase} set containing words and phrases that are likely to form an organization when concatenated to a location, such as ``\emph{Bank of}'' (before America). Such phrases form a set of size 82. We then perform eligibility check by locating those organization entities containing one of such phrases and change their entity type by deleting those phrases (e.g. delete ``\emph{Re}'' from ``\emph{Munich Re}'' ). \textbf{Entity Context Change} involves the insertion of a natural context after the entity, such as ``\emph{'s largest corporation}'' and ``\emph{'s football club}''. We have 16 of such contexts.
\paragraph{Person}
%Similar to transitioning to \textsc{Organization}, the augmented entity for transitioning to \textsc{Person} is not restricted to mainstream names such as Austin (in Texas US) or Charlotte (in North Carolina US). %For example, for the transition to a \textsc{Person} type from a \textsc{Location}, there are only a couple locations that people accept as mainstream names such as Austin (in Texas US), Charlotte (in North Carolina US), and Brooklyn (borough of New York US). 
%Given the example original text in Table \ref{table:heuristic_rule_based_augmentation} for the transition to \textsc{Person}, the single token entity ``\emph{Colts}'' is identified as an eligible entity because of the overlapping token between the eligible organization entity and potential person entity in this case is ``\emph{Colts}''. We constructed word  ``\emph{phrase}'' sets corresponding to \textsc{Person} entity changes consisting of 152 word ``\emph{phrases}'' and context changes consisting of 49 word ``\emph{phrases}'' via scripts and manual data collection.  ``\emph{Zardari}'' is sampled from the  word  ``\emph{phrase}'' set for \textsc{Person} entity changes and is inserted after ``\emph{Colts}''. ``\emph{and her team}'' is then sampled from the  word  ``\emph{phrase}'' set for \textsc{Person} context changes and is inserted after ``\emph{Zardari}'' to maintain semantic meaning. These augmentations cause the entity to transition from an \textsc{Organization} to a \textsc{Person}.
Similar to transitioning to \textsc{Organization}, an example is considered \textbf{eligible} for transitioning to \textsc{Person} if an entity only contains one token (e.g. ``\emph{Colts}''). \textbf{Entity Token Change} in this situation refers to the insertion of a token representing a person's last name after the original token to change the entity type to  \textsc{Person} (e.g. add ``\emph{Zardari}'' after ``\emph{Colts}''). Such predefined tokens for insertion form a set of size 152, including examples such as ``\emph{Dutra}'' and ``\emph{Martin}''. \textbf{Entity Context Change} for a person then involves inserting a suitable context after the newly formed entity, such as ``\emph{and her team}'' and ``\emph{and his company}''. Such phrases form a set of size 49.

We include more examples of word phrases in the Appendix (Table \ref{table:word_phrases}) and the GitHub repository contains the full sets. Note that the automatically augmented adversarial examples may lack semantic and syntactic quality. For example, there may be grammatical issues or the randomly inserted contexts may be in conflict with current contexts. Thus we only use them for adversarial training (Section \ref{sec: Mixup}). To build the challenging test set, we manually select the high quality examples from the augmented test dataset (Section \ref{sec:challengeset}).

% The transition to the \textsc{Location} label is different in the sense that we restrict the set of possible augmented entities to only  contain entities that are real world locations and is further explained in Appendix \ref{appendix:Location}.
%For transitioning to \textsc{Organization} and \textsc{Person}, an example is considered eligible for the heuristic augmentation if an entity contains within it only a single token and its augmented version does not have to exist in the real world. The transition to the \textsc{Location} type is different in that it is restricted to the set of possible augmented entities that are real world locations.%
%\diyi{please add necessary details rather than always referring people to appendix}  
%%%%%%%%\subsection{Step 2: Performing Expert-Guided Adversarial Augmentation}

\section{Mixup with Adversarial Examples}
\label{sec: Mixup}
Adversarial training improves a model's robustness to adversarial examples by directly training on adversarial examples, however, such training might hurt generalization \citep{raghunathan2019adversarial} or cause overfitting on adversarial features \citep{lee2020adversarial} (predefined \emph{word phrases} in our case).
To this end, we leverage mixup \cite{zhang2018mixup, verma2019manifold} to mitigate these issues and further improve generalization on the basis of adversarial training \citep{lee2020adversarial}. 

Given a pair of data points $(x,y)$ and $(x^{\prime},y^{\prime})$, where $x$ denotes a data point and $y$ denotes its label in a one-hot representation, mixup \citep{zhang2018mixup} creates a new data point by the interpolation of the data and their labels as shown below with $\lambda$ being drawn from a beta distribution:
\begin{equation} \hat{x}= \lambda x +(1 - \lambda) x^{\prime}  \end{equation}
\begin{equation} \hat{y}= \lambda y +(1 - \lambda) y^{\prime}  \end{equation}

% The mixup technique trains the neural network for image classification by the minimization of the loss on the virtual examples. The linear interpolation of the data points creates a virtual vicinity distribution around the original data space causing the improvement of the classifier's generalization performance when being trained on the interpolated data points.

In this work, $(x,y)$ is a training example that is eligible for heuristic augmentation and is paired with its heuristically modified version $(x^{\prime},y^{\prime})$. Since textual data is discrete and cannot be mixed in the input space, the interpolation of the two examples is computed in the hidden space. 

Following \citet{chen2020local}, Let $\mathbf{h}^m = \{ h_{1} ..h_{n} \}$ be the hidden representations after the $m$-th layer where they are the concatenation of the token representations. The hidden representation for each token in the original example at the $m$-th layer $\mathbf{h}^m$ is linearly interpolated with $\mathbf{h}^{m\prime}$, the representation for each token in the augmented example, by a ratio $\lambda$:

\begin{equation} \mathbf{\hat{h}}^m = \lambda \mathbf{h}^m + (1-\lambda) \mathbf{h}^{m\prime} \end{equation}

Then $\mathbf{\hat{h}}^m$ is passed to the $(m + 1)$-th layer, and the labels for the final output logits are mixed at the same ratio. $m$ is randomly sampled from $\{ 8,9,10 \}$. The mixing parameter $\lambda$ is sampled from a beta distribution: $ \lambda \sim \mathit{B}(\alpha,\beta) $, where $\alpha$ and $\beta$ determine the skew of the beta distribution. In this work, we use two different beta distributions from which to sample $\lambda$. For each pair of data points, two mixed data points are generated. One data point is closer to the original examples and the other is closer to the adversarial examples.
%One beta distribution for when the original examples are to be mixed with their heuristically augmented versions, and another for when those heuristically augmented examples are to be mixed with their original counterparts.
See Appendix \ref{appendix:mixuptuning} for more details.

\section{Experiments}
\begin{table*}[ht]
\centering
% \begin{tabular}{p{0.1\linewidth} | p{0.22\linewidth} | p{0.18\linewidth} | p{0.21\linewidth} | p{0.16\linewidth}}
\begin{tabular}{c|l|c|c|c}
\hline
\textbf{Percent} & \textbf{Model} & \textbf{ID} & \textbf{CS} & \textbf{OOD}\\
\hline\cline{1-5}
\textbf{N/A} & BERT &          90.82 &    71.80      &58.72\\ 
\hline\cline{1-5}
\textbf{N/A} & BERT + TAVAT &   91.82  &   70.14    & - \\
%0.3295  
\hline\cline{1-5}
\textbf{10\%} & BERT + AT & 90.37 & 86.16 &61.09 \\ 
\cline{2-5}
& BERT + AT + Dropout & 90.1 & 84.97&  61.86 \\ 
\cline{2-5}
& BERT + AT + Mixup & \textbf{90.79} & \textbf{88.79} &  \textbf{67.47}\\
\cline{2-5}
& BERT + TextFlint & 88.85 &54.04   &66.67 \\
\hline\cline{2-5}
\textbf{30\%} & BERT + AT & 90.84 &86.42  &60.76\\
\cline{2-5}
& BERT + AT + Dropout & \textbf{90.93} & 86.91  &61.6 \\ 
\cline{2-5}
& BERT + AT + Mixup & 90.85 & \textbf{87.30}   &\textbf{69.46}\\ 
\cline{2-5}
& BERT + TextFlint & 89.71 & 60.32 & 65.88\\
\hline\cline{2-5}
\textbf{50\%} & BERT + AT & 90.85 &87.50 &62.18 \\ 
\cline{2-5}
& BERT + AT + Dropout & 90.19 & \textbf{88.88}  &60.83 \\ 
\cline{2-5}
& BERT + AT + Mixup &\textbf{90.92} & 88.00 &\textbf{67.47} \\ 
\cline{2-5}
& BERT + TextFlint & 89.55 & 53.49  &65.48 \\
\hline\cline{2-5}
\textbf{100\%} & BERT + AT & 90.52 &  87.74  &57.76 \\ 
\cline{2-5}
& BERT + AT + Dropout &90.16  & 88.45  &60.25 \\ 
\cline{2-5}
& BERT + AT + Mixup & \textbf{90.53} & \textbf{90.21}  & 67.07\\ 
\cline{2-5}
& BERT + TextFlint &87.31  &59.12  & \textbf{69.05}\\
\hline
\end{tabular}
\caption{\small{F1 Scores on the original CoNLL 2003 Test Set (ID), proposed Challenging Set (CS), and Out of Domain Test Set (OOD). All the results were averaged over 3 runs. `-' refers to unstable training which causes the model to collapse. Note that in the third and fourth columns, models are trained on CoNLL 2003 training data (and their augmented versions if adversarial training is available). In the fifth column, models are trained on CoNLL 2003 training data and 5-shot examples from the OntoNotes training data (and their augmented versions if adversarial training is available).}} 
\label{table:f1_main_results}
\end{table*}

\subsection{Datasets and Pre-processing}
\label{sec:challengeset}
\paragraph{In-Distribution dataset}(ID) We use CoNLL 2003 \cite{tjong-kim-sang-de-meulder-2003-introduction} with the BIO labeling scheme following \citet{chen2020local}. In order to make mixup possible in recent transformer based models like BERT, we assigned labels to the special tokens [SEP], [CLS], and [PAD]. All models are trained on the ID training set by default. We report the results on the ID test set in the third column of Table \ref{table:f1_main_results}.

\paragraph{Challenge Set}(CS) For the challenging set, two graduate students who have linguistic backgrounds and are familiar with NER tasks, manually constructed the dataset consisting of the ID test set transformed by the expert-guided augmentations.
The goal was to build a challenging test set containing only high quality data points, by manually labeling the quality (as high or low) and making small corrections.
Before annotating the full set of augmented data, they did a test annotation of a sample size of 50 examples to calculate the annotator agreement  and the resulting annotator agreement was 78\%.
They then manually annotated the full augmented test set which resulted in a challenging set of 1000 high quality data points. 
% During the annotation and cleaning of this test set, we made small corrections to ensure grammatical correctness. 
 
% We exclude 25\% of the word phrases that are used in the construction of the CS from the word phrase sets used in the constructing of the adversarial training examples.

\paragraph{Out-of-Domain}(OOD) In addition to training on an ID training set and testing on an ID test set and challenging set, we further test the few-shot generalization ability of our proposed approach on an out-of-domain dataset: OntoNotes \citep{inbook}. In this setting, all models are given 5 training examples of each class from the OntoNotes \citep{inbook} training set (along with the ID training data). After training, we tested their out-of-domain generalization by using an OOD test set consisting of 50 examples from the OntoNotes test set. All data points had to follow the condition that the percentage of entity tokens out of all tokens is greater than 49\%. This condition serves the purpose of allowing for the evaluation of the model’s performance upon mostly entity tokens. Note that OntoNotes has a more fine-grained entity category than CoNLL 2003, so we mapped the OntoNotes labels to the CoNLL 2003 labels so that the data would be compatible with our models. %(see Appendix \ref{appendix:generationcode}).

\subsection{Baselines and Model Settings}

We train six types of models: (1) a BERT Base \cite{devlin-etal-2019-bert} model on only the original training examples  (\textsl{BERT}); (2) a BERT Base model on the original training examples and training examples that are augmented with the expert-guided adversarial heuristics (\textsl{BERT+AT}); (3) a \textsl{BERT+AT} model with dropout probability of 0.5 \cite{hinton2012improving} (\textsl{BERT + AT + Dropout}); (4) a BERT Base model utilizing Token-Aware Virtual Adversarial Training (TAVAT, \citealp{li2020tavat}), a gradient-based adversarial training technique  (\textsl{BERT + TAVAT});
(5) a BERT Base model trained with the text-based adversarial attacks proposed in \citet{gui2021textflint} utilizing their defined NER transformations (Appendix \ref{appendix:textflint})  (\textsl{BERT + TextFlint});
(6) a BERT Base model utilizing mixup to linearly interpolate the original training examples with the expert-guided adversarial examples (\textsl{BERT + AT + Mixup}).
Note that models using mixup are not trained on more data points, since two mixed data points are generated given a pair of data points (see Section \ref{sec: Mixup}).
% In order to allow for all non-mixup models utilizing text-based adversarial training to be trained on the same number of datapoints as \textsl{BERT+AT+Mixup}: Let the number of eligible examples be $n$ and the number of ineligible examples be $m$.  The number of data points to be trained on equals $2*n + m$.

In order to test the generalization ability of the models using the proposed adversarial augmentation, we varied the percentage of adversarial augmented examples (10\%, 30\%, 50\%, and 100\% of the total number of eligible examples) used for both the proposed adversarial training and TextFlint \citep{gui2021textflint}.
We also used smaller predefined \emph{word phrase} sets to augment the training data by excluding 25\% of the total word phrases used in the construction of the CS.

% corresponding to 553, 1,657, 2761, 5523 examples respectively). 
% 5-shot training is used to provide limited OOD data during training to then be evaluated on the OOD test set. The support set is created by sampling 5 examples containing each class as in \cite {huang2020fewshot}. The augmented version of each example in the support set are included in the training for the models that utilize text-based adversarial training.

\subsection{Results and Analysis}
%\diyi{please carefully describe the results by situating it to the key claims of this work}
\paragraph{CS} As shown in Table \ref{table:f1_main_results}, \textsl{BERT} had a significant performance decline when tested on the CS, and the prior adversarial training approach failed to increase the performance on CS, demonstrating the novel challenge proposed. Not surprisingly, \textsl{BERT+AT} can dramatically improve the model's performance on the CS, even when only 10\% of the eligible augmentation is used. Incorporating mixup can consistently improve it as demonstrated on CS. While prior adversarial training severely hurt the model's performance on ID, \textsl{BERT+AT+Mixup} almost maintained its ID performance which suggests the good generalization ability training with the proposed adversarial augmentation provides.
% While training on TextFlint examples increases OOD performance, it causes a severe degradation to BERT's performance on ID and CS, which could be due to the noise injected by random swapping of entities.

For an ablation study, we conducted experiments in which we used mixup to interpolate pairs of ID training data points, and observed a big performance gap when compared to our approach (see Figure \ref{figure:ablationtable} in Appendix). This proved the strategic design of mixing original examples and their expert-guided adversarial versions.

\paragraph{OOD} In the few-shot generalization experiments, while the original \textsl{BERT} demonstrated poor performance on OOD, TextFlint significantly increased performance. \textsl{BERT + AT} only marginally outperforms \textsl{BERT} when limited examples are augmented, probably suggesting that the lack of generalization is due to naive adversarial training on the proposed augmentation. However, \textsl{BERT+AT+Mixup} significantly increased the performance as demonstrated by achieving the best performance (69.46), while also outperforming the baselines in most settings.
Other than the learning of smoother decision boundaries, we also hypothesize that the interpolated representations enhance the quality of the adversarial examples' representations, thus resulting in improved generalization. This hypothesis is based on the fact that the quality of the augmented examples is sometimes limited. So the interpolation with the original data in the hidden space may help to improve the quality.
\section{Conclusion}
This work proposed an expert-guided adversarial augmentation for NER consisting of the altering of entity types by strategic selection and modification of tokens and their contexts. Using this augmentation strategy on CoNLL 2003 and manually filtering the generated examples for quality, we constructed a high-quality challenging test set for the NER task.
%created using expert-guided heuristics to facilitate the testing of out-of-distribution performance of NER systems. 
We show that SOTA NER systems suffer from dramatic performance drop when evaluated on our challenging set.
Beyond simply using the proposed augmentation for adversarial training, we demonstrated that leveraging mixup between original examples and their augmented versions can outperform state-of-the-art baselines on in-distribution data, the challenging set, and few-shot generalization to out-of-domain data.%\diyi{if you already solve this using your technique, what's the value of the dataset?} 

\section*{Acknowledgment}
We would like to thank the anonymous reviewers
for their helpful comments, and the members of the Georgia Tech SALT lab for their feedback.

\bibliographystyle{acl_natbib}
\bibliography{ref}

\appendix
\section{Expert-Guided Augmentation's Adversarial Properties}
\label{appendix:AdversarialProperties}
When the expert-guided augmentation is applied to an example, the entity’s new label is now the ground truth label. If the model classifies based upon the spurious correlation between the remnants of the original entity and context with the original label within the newly augmented text, it will be provoking the wrong classification by the prediction of the old label. This demonstrates the augmented example’s adversarial properties.

%\subsection{Transition to Location Type}
%\label{appendix:Location}
%In order to ensure the augmented entity of type \textsc{Location} is a real world location, the organization entities that have overlapping tokens with location entities need to be identified. For this we constructed word  ``\emph{phrase}'' sets corresponding to \textsc{Location} entity changes consisting of 82 word ``\emph{phrases}'' and context changes consisting of 16 word ``\emph{phrases}'' via scripts and manual data collection. Given the example original text in Table \ref{table:heuristic_rule_based_augmentation} for the transition to \textsc{Location}, the single token entity ``\emph{Munich}'' is identified as an eligible entity because of the overlapping token between the eligible organization entity and potential location entity is in this case is ``\emph{Munich}''. ``\emph{Re}'' is identified as located within the  word  ``\emph{phrase}'' set for \textsc{Location} entity changes and is removed from the text causing the entity ``\emph{Munich}'' to transition to a \textsc{Location}. ``\emph{'s largest corporation}'' is inserted after the augmented entity to maintain semantic quality.

\section{ Mixup Implementation Details and Hyperparameter Tuning}
\label{appendix:mixuptuning}
After sampling a $\lambda$ from the beta distribution, we modify it by applying $\lambda = max (\lambda, 1 - \lambda)$, which guarantees that the $\lambda$ to be used is no less than $0.5$. A large $\lambda$ can guarantee that the resulting mixed data point ($\hat{x}= \lambda x +(1 - \lambda) x^{\prime}$) is always closer to $x$.
We use two different beta distributions to sample the mixing parameter from, one for when the original examples are to be mixed (original examples as $x$, augmented examples as $x^{\prime}$) and one for when the heuristically augmented examples are to be mixed (augmented examples as $x$, original examples as $x^{\prime}$). 

For the two hyperparameters corresponding to each of the two beta distributions from which the mixing parameter is sampled, $\alpha$ and $\beta$, we first set them at 200 and 5 respectively. We experimented with lessening the skew of the beta distribution decreasing $\alpha$ to 150 and while keeping $\beta$ at 5. We then further experimented with increasing its skew by decreasing $\alpha$ to 130 and while at the same time increasing $\beta$  to values of 7 and 9. 

In the few-shot generalization experiments, our implementation of mixup uses four different beta distributions from which to sample the mixing parameter: Similarly, two for the in-distribution original and augmented training examples, and two for the out-of-domain original and augmented training examples.  
%one beta distribution for when the in-distribution examples are to be mixed with their heuristically augmented versions, and another for when those heuristically augmented examples are to be mixed with their in-distribution counterparts. There is also a beta distribution for when the out-of-domain examples are to be mixed with their heuristically augmented versions and another for when those heuristically augmented examples are to be mixed with their out-of-domain counterparts. The values of the shape parameters used in the tuning of these beta distributions are similar to the ones used during training without 5-shot.

% \section{Adversarial Training Set Generation Code}
% \label{appendix:generationcode}
% The guided-adversarial training examples used in the experiments are generated using code with the expert-guided heuristic word “phrase” sets given as input. The code proposed in this work can also be easily repurposed for other NER datasets due to its linguistic universality with the expert-guided heuristic word “phrase” sets for other domains being given as input to the code. 

\begin{figure}[t]
    \centering
    \includegraphics[width=0.9\linewidth]{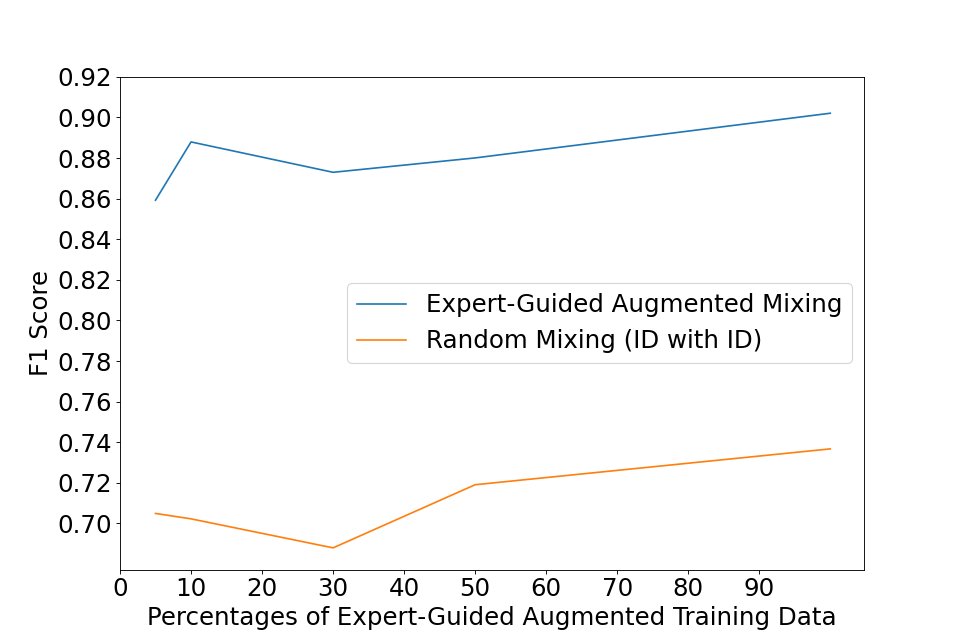}
    \caption{Random Mixing of \textbf{ID data} with \textbf{ID data} vs. Mixing of \textbf{ID data} with \textbf{Expert-Guided Augmented data}; Performances are on the CS} 
    \label{figure:ablationtable}
\end{figure}

\begin{table*}[ht]
\label{appendix:noheldouttable}
\centering
\begin{tabular}{c|l|c}
\hline
\textbf{Percent} & \textbf{Model} &  \textbf{Challenge Set}\\
\hline\cline{1-3}
\textbf{10\%} & BERT + AT   &88.53   \\ 
\cline{2-3}
& BERT + AT + Dropout   &83.98   \\ 
\cline{2-3}
& BERT + AT + Mixup   &88.54   \\
\hline\cline{2-3}
\textbf{30\%} & BERT + AT  &91.16  \\
\cline{2-3}
& BERT + AT + Dropout   &93.08  \\ 
\cline{2-3}
& BERT + AT + Mixup  &93.09   \\ 
\hline\cline{2-3}
\textbf{50\%} & BERT + AT   & 88.74  \\ 
\cline{2-3}
& BERT + AT + Dropout   &93.38   \\ 
\cline{2-3}
& BERT + AT + Mixup  &92.48   \\ 
\hline\cline{2-3}
\textbf{100\%} & BERT + AT   &92.97   \\ 
\cline{2-3}
& BERT + AT + Dropout  &93.77  \\ 
\cline{2-3}
& BERT + AT + Mixup   &92.33  \\ 
\hline
\end{tabular}
\caption{F1 scores on the challenging set when no \emph{word phrases} were held out during training; All of the results were averaged over 3 runs.}
\label{table:noheldouttable}
\end{table*}
%\section{5-Shot Experimental Setting}
%\label{appendix:fewshotsettings}
%For the 5-shot setting, the support set is created by sampling 5 examples containing each class as in \cite {huang2020fewshot}. The augmented version of each example in the support set are included in the training for the models that utilize text-based adversarial training.%

\section{TextFlint NER Task Specific Transformations}
\label{appendix:textflint}
The four TextFlint NER task specific transformations used are ConcatSent, EntTypos, CrossCategory, and SwapLonger. ConcatSent involves the concatenation of two sentences into a longer one. EntTypos involves the swapping/deleting/adding of a random character to entities. CrossCategory involves the swapping of entities with ones that can be labeled by different labels. SwapLonger involves the substituting of the short entities for longer ones. Since only ConcatSent and EntTypos were available through the TextFlint framework during the time of this work, we reimplemented CrossCategory and SwapLonger for the experiments.

\section{No Word Phrases Held Out Experiments}
\label{appendix:noheldout}

In Table \ref{table:noheldouttable}, we provide the results when using all of the \emph{word phrases} for adversarial augmentation during training. Compared to the setting where 25\% of the word phrases were held out for training (Table \ref{table:f1_main_results}), the models experienced a significant drop in performance. The models may have learned the spurious correlation between the words from the \emph{word phrase} set and the entity labels instead of learning the linguistic relation.
This demonstrates that even though BERT's performance increases when trained on the expert-guided augmented data, the challenging set is still not "solved" as the removal of 25\% of the word phrases from training caused this significant of a performance drop. This ``held out'' setting simulates the real world deployment of NER models.

\section{Tuning of TAVAT's Hyperparameters}
The hyperparameters unique to Token-Aware Virtual Adversarial Training (TAVAT) such as the adversarial training step, the constraint bound of the pertubation, the adversarial step size, and the initialization bound are tuned using the values in \citet{li2020tavat}.

\begin{table*}[]
\small
\begin{tabular}{l|l|l}
\toprule
Target Entity & Word Phrase Set       & Examples                                                                                 \\ \midrule
Organization  & Entity Token Change   & Department of Transportation | Reserve Bank of | Workers Party | Corporation             \\
              & Entity Context Change & , and its ministers, | 's star player | and its services | with its government officials \\ \midrule
Location      & Entity Token Change   & Court of Appeals | Stock Exchange | UNITED | Radio                                       \\
              & Entity Context Change & 's leading newsroom  | 's countryside | 's hockey team                                   \\ \midrule
Person        & Entity Token Change   & Doorn | Liano | Bronckhorst | Aynaoui | Goey  | Sidhu | Bedie                            \\
              & Entity Context Change & 's company | and other politicians | , an accomplished player                            \\ \bottomrule
\end{tabular}
\caption{More examples from the predefined \emph{word phrase} sets ; A vertical bar ( | ) is used to separate word phrases.}
\label{table:word_phrases}
\end{table*}

\section{Experimental Details:}

\subsection{Description of computing infrastructure used:}
\noindent
GEFORCE RTX 2080    CUDA Version: 11.0

\subsection{Runtime} 
\begin{itemize}
    \item Training: 2 to 2 and 1/2 hours.
    \item Inference: 3 minutes or less
\end{itemize}

\subsection{Parameters}
\noindent
BERT contains 110 million parameters.

\subsection{Hyperparameters for Training without 5-Shot}
\begin{itemize}

\item BERT: max sequence length 256,  batch size 8,  number of training epochs 10, adam epsilon=1e-08, learning rate=5e-05, weight decay=0.0

\item All dropout models have dropout probability set to 0.5 for all fully connected layers in the embeddings, encoder, and pooler.

\item Mixup 10 \% Augmented data:
\begin{itemize}
    \item Original examples: $\alpha$=130 $\beta$=9
    \item Augmented examples: $\alpha$=200 $\beta$=5
  \end{itemize}

\item Mixup 30 \% Augmented data: 
\begin{itemize}
    \item Original examples:  $\alpha$=150 $\beta$=5
    \item Augmented examples: $\alpha$=200 $\beta$=5
  \end{itemize}

\item Mixup 50 \% Augmented data: 
\begin{itemize}
    \item Original examples:  $\alpha$=130 $\beta$=7
    \item Augmented examples: $\alpha$=200 $\beta$=5
  \end{itemize}
  
\item Mixup 100 \% Augmented data:
\begin{itemize}
    \item Original examples:  $\alpha$=150 $\beta$=5
    \item Augmented examples: $\alpha$=200 $\beta$=5
  \end{itemize}

\item TAVAT Model:
adv init mag=0.2, adv lr=0.05, adv max norm=0.5, adv steps=2, adv train=1
\end{itemize}
\begin{itemize}
\subsection{Hyperparameters for 5-Shot Training}
\item Mixup 10 \% Augmented data:
\begin{itemize}
    \item Original examples: $\alpha$=150 $\beta$=5
    \item Augmented examples: $\alpha$=200 $\beta$=5
    \item Original OOD examples: $\alpha$=200 $\beta$=5
    \item Augmented OOD examples: $\alpha$=130 $\beta$=7
  \end{itemize}

\item Mixup 30 \% Augmented data:
\begin{itemize}
    \item Original examples: $\alpha$=200 $\beta$=5
    \item Augmented examples: $\alpha$=150 $\beta$=5
    \item Original OOD examples: $\alpha$=200 $\beta$=5
    \item Augmented OOD examples: $\alpha$=130 $\beta$=7
  \end{itemize}

\item Mixup 50 \% Augmented data:
\begin{itemize}
    \item Original examples: $\alpha$=150 $\beta$=5
    \item Augmented examples: $\alpha$=200 $\beta$=5
    \item Original OOD examples: $\alpha$=200 $\beta$=5
    \item Augmented OOD examples: $\alpha$=130 $\beta$=7
  \end{itemize}

\item Mixup 100  \% Augmented data:
\begin{itemize}
    \item Original examples: $\alpha$=130 $\beta$=5
    \item Augmented examples: $\alpha$=200 $\beta$=5
    \item Original OOD examples: $\alpha$=200 $\beta$=5
    \item Augmented OOD examples: $\alpha$=130 $\beta$=7
  \end{itemize}
  
\item TAVAT Model, 5-Shot Training: adv init mag=0.2, adv lr=0.05, adv max norm=0.5, adv steps=2, adv train=1
\end{itemize}

\subsection{Dataset}
\begin{itemize}
\item CoNLL 2003 Language: English

\item Training set for CoNLL 2003: Number of examples: 14041

\item Dev set for CoNLL 2003: Number of examples: 3250

\item Test set for CoNLL 2003: Number of examples: 3453

\end{itemize}

\end{document}